%% file: samplepaper.tex
\documentclass[runningheads]{llncs}
\usepackage{graphicx}
\usepackage{amsmath,amssymb} 
\usepackage{color}
\usepackage{overpic}
\usepackage{microtype}

\usepackage{tikz}
\usepackage{pgfplots}
\usetikzlibrary{calc,trees,positioning,arrows,chains,shapes.geometric,%
    decorations.pathreplacing,decorations.pathmorphing,shapes,%
    matrix,shapes.symbols}

\begin{document}
\title{DeepWrinkles: Accurate and Realistic Clothing Modeling}
\titlerunning{DeepWrinkles: Accurate and Realistic Clothing Modeling}
%
\author{Zorah L\"{a}hner \inst{1,2} \and
Daniel Cremers\inst{2} \and
Tony Tung\inst{1}}
\authorrunning{Z. L\"{a}hner et al.}
%
\institute{Facebook Reality Labs, USA, \email{tony.tung@fb.com}\\ \and
Technical University Munich, Germany, \email{\{laehner,cremers\}@in.tum.de}}
\maketitle              

\begin{figure}[t]
	\includegraphics[trim={9cm 4cm 8cm 2.2cm},clip,width=.195\linewidth]{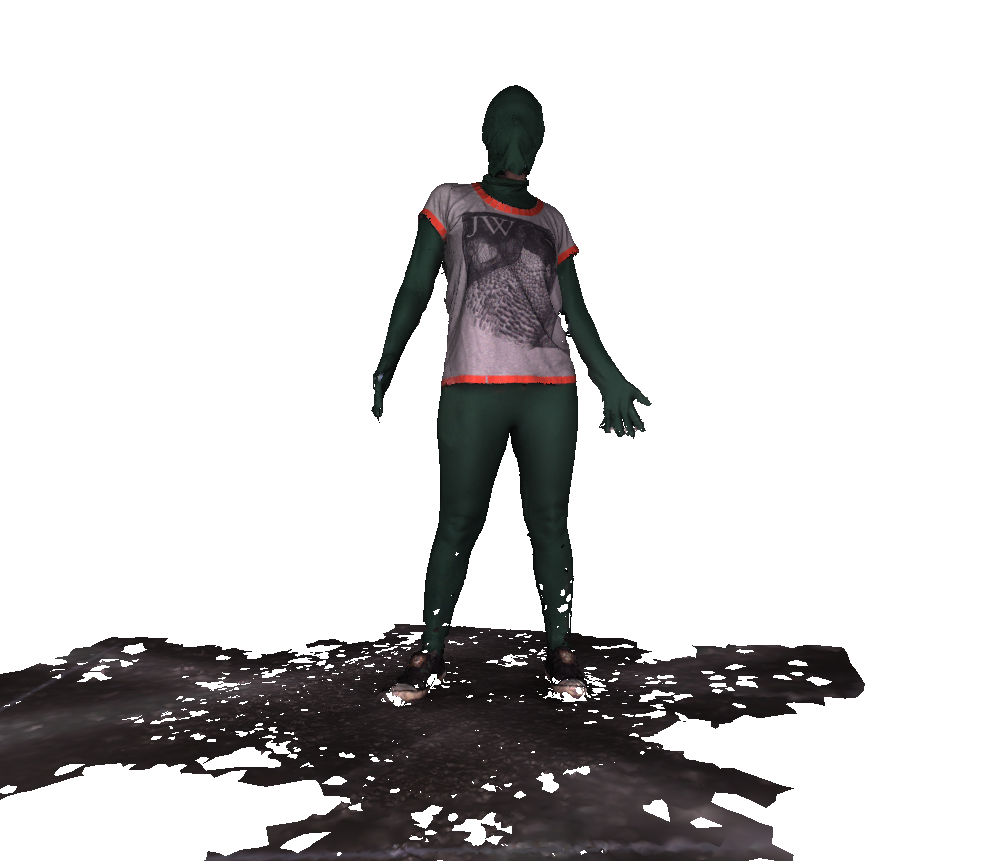}
    \quad
    \includegraphics[trim={12cm -1cm 10cm 0cm},clip,width=.21\linewidth]{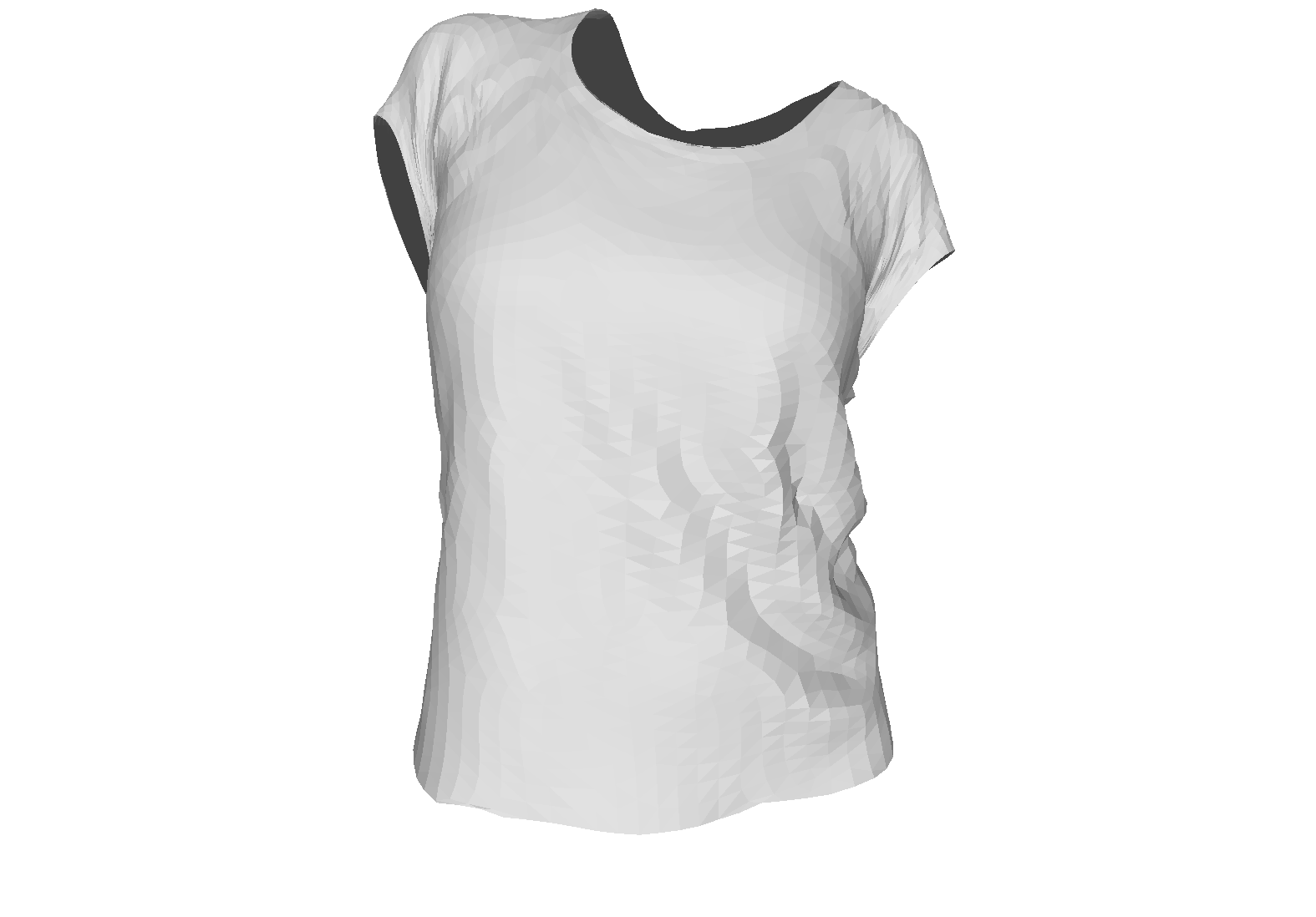}
    \quad
    \includegraphics[trim={0cm -3cm 0cm 00cm},clip,width=.26\linewidth]{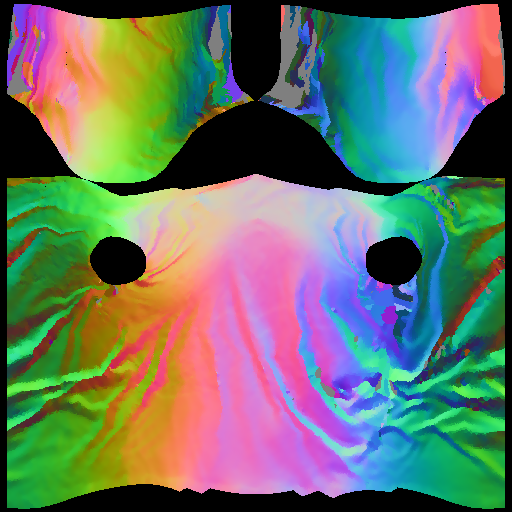}
    \qquad
    \includegraphics[trim={2cm 1.3cm 1cm 0.5cm},clip,width=.155\linewidth]{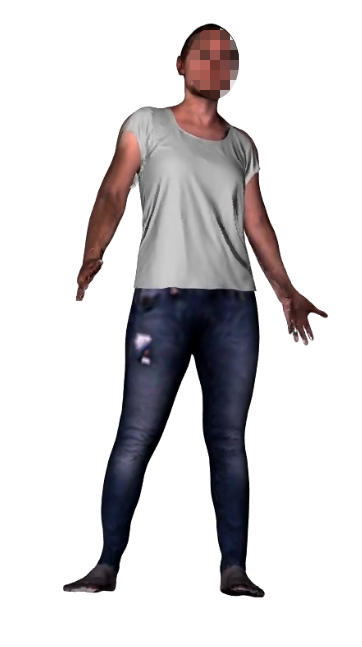}
    \caption{Accurate and realistic clothing modeling with DeepWrinkles, our entirely data-driven framework. (Left) 4D data capture. (Middle Left) Reconstruction from subspace model. (Middle Right) Fine wrinkles in normal map generated by our adversarial neural network. (Right) 3D rendering and animation on virtual human.}
    \label{fig:overview}
\end{figure}

\begin{abstract}
We present a novel method to generate accurate and realistic clothing deformation from real data capture.
Previous methods for realistic cloth modeling mainly rely on intensive computation of physics-based simulation (with numerous heuristic parameters), while models reconstructed from visual observations typically suffer from lack of geometric details.
Here, we propose an original framework consisting of two modules that work jointly to represent global shape deformation as well as surface details with high fidelity.
Global shape deformations are recovered from a subspace model learned from 3D data of clothed people in motion, while 
high frequency details are added to normal maps created using a conditional Generative Adversarial Network whose architecture is designed to enforce realism and temporal consistency.
This leads to unprecedented high-quality rendering of clothing deformation sequences, where fine wrinkles from (real) high resolution observations can be recovered.
In addition, as the model is learned independently from body shape and pose, the framework is suitable for applications that require retargeting (e.g., body animation).
Our experiments show original high quality results with a flexible model. We claim an entirely data-driven approach to realistic cloth wrinkle generation is possible.

\end{abstract}
\section{Introduction}\label{sec:introduction}
Realistic garment reconstruction is notoriously a complex problem and its importance is undeniable in many research work and applications, such as accurate body shape and pose estimation in the wild (i.e., from observations of clothed humans), realistic AR/VR experience, movies, video games, virtual try-on, etc.

For the past decades, physics-based simulations have been setting the standard in movie and video game industries, even though they require hours of labor by experts.
More recently methods for full clothing reconstruction using multi-view videos or 3D scan systems have also been proposed \cite{ClothCap17}. Global deformations can be reconstructed with high fidelity semi-automatically. Nevertheless, accurately recovering geometric details such as fine cloth wrinkles has remained a challenge.

In this paper, we present DeepWrinkles (see Fig.~\ref{fig:overview}), a novel framework to generate accurate and realistic clothing deformation from real data capture.
It consists of two complementary modules: (1) A statistical model is learned from 3D scans of clothed people in motion, from which clothing templates are precisely non-rigidly aligned.
Clothing shape deformations are therefore modeled using a linear subspace model, where human body shape and pose are factored out, hence enabling body retargeting.
(2) Fine geometric details are added to normal maps generated using a conditional adversarial network whose architecture is designed to enforce realism and temporal consistency.

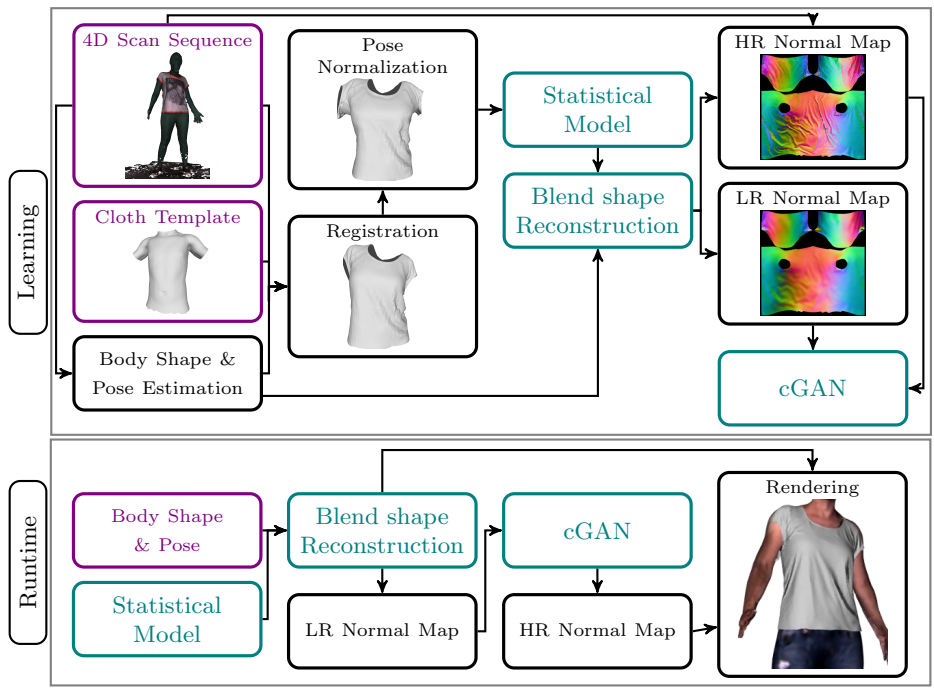
\begin{figure}[t]
	\input{overview2.tikz}
    \caption{Outline of DeepWrinkles. 
    During Learning, we learn to reconstruct global shape deformations using a Statistical Model and a mapping from pose parameters to Blend shape parameters  using real-world data. We also train a neural network (cGAN) to generate fine details on normal maps from lower resolution ones.
    During runtime, the learned models reconstruct shape and geometric details given a priori body shape and pose. Inputs are in violet, learned models are in cyan. 
    }
    \label{fig:diagram}
\end{figure}

To our knowledge, this is the first method that tackles 3D surface geometry refinement using deep neural network on normal maps.
With DeepWrinkles, we obtain unprecedented high-quality rendering of clothing deformation, where global shape as well as fine wrinkles from (real) high resolution observations can be recovered, using an entirely data-driven approach.
Figure~\ref{fig:diagram} gives an overview of our framework with a T-shirt as example.
Additional materials contain videos of results. We show how the model can be applied to virtual human animation, with body shape and pose retargeting.

\section{Related Work}\label{sec:relatedwork}

Cloth modeling and garment simulation have a long history that dates back to the mid 80s.
A general overview of fundamental methods is given in \cite{DBLP:books/daglib/0007582}.
There are two mostly opposing approaches to this problem. One is using physics-based simulations to generate realistic wrinkles, and the other captures and reconstructs details from real-world data.

\subsubsection{Physics-based simulation.}\label{subsec:clothsimulation}
For the past decades, models relying on Newtonian physics have been widely applied to 
simulate cloth behavior. They usually model various material properties such as stretch (tension), stiffness, and weight. For certain types of applications (e.g., involving human body) additional models or external forces have to be taken into account such as body kinematics, body surface friction, interpenetration, etc~\cite{Choi:2002:SBR:566570.566624,DBLP:conf/sca/BridsonMF03,Baraff:2003:UC:882262.882357,DBLP:journals/tog/GoldenthalHFBG07}.
Note that several models have been integrated in commercial solutions (e.g., Unreal Engine APEX Cloth/Nvidia NvCloth, Unity Cloth, Maya nCloth, MarvelousDesigner, OptiTex, etc.)~\cite{MarvelousDesigner}. 
Nevertheless, it typically requires hours or days if not weeks of computation, retouching work, and parameter tuning by experts to obtain realistic cloth deformation effects.

\subsubsection{3D capture and reconstruction.}\label{subsec:clothestimatio}
Vision-based approaches have explored ways to capture cloth surface deformation under stress, and estimate material properties through visual observations for simulation purpose~\cite{Wang:2011:DDE,DBLP:journals/cgf/MiguelBTBMOM12,DBLP:books/daglib/0029461,DBLP:conf/iccv/LeroyFB17}.
As well, several methods directly reconstruct whole object surface from real-world measurements. \cite{DBLP:journals/tog/WhiteCF07} uses texture patterns to track and reconstruct garment from a video, 
while 3D reconstruction can also be obtained from multi-view videos without markers~\cite{DBLP:conf/iccv/TungNM09,DBLP:books/daglib/0029461,DBLP:conf/iccv/LeroyFB17,DBLP:journals/cga/StarckH07,Vlasic:2008:AMA:1399504.1360696,Bradley:2008:MGC:1360612.1360698}.
However without sufficient prior, reconstructed geometry can be quite crude.
When the target is known (e.g., clothing type), templates can improve the reconstruction quality \cite{DBLP:journals/tog/AguiarSTAST08}.
Also, more details can be recovered by applying super-resolution techniques on input images \cite{DBLP:conf/cvpr/TungNM08,DBLP:conf/iccv/GoldluckeC09,DBLP:conf/cvpr/TsiminakiFB14,Bradley:2008:MGC:1360612.1360698}, or using photometric stereo and information about lighting \cite{DBLP:conf/iccv/HernandezVBSC07,Vlasic:2009:DSC:1661412.1618520}.
Naturally, depth information can lead to further improvement~\cite{DBLP:conf/cvpr/NewcombeFS15,Dou:2016:FRP:2897824.2925969}.

In recent work~\cite{ClothCap17}, cloth reconstruction is obtained by clothing segmentation and template registration from 4D scan data. Captured garments can be retargeted to different body shapes.
However the method has limitations regarding fine wrinkles.

\subsubsection{Coarse-to-fine approaches.}\label{subsec:datadriven}
To reconstruct fine details, and consequently handle the bump in resolution at runtime (i.e., higher resolution meshes or more particles for simulation), methods based on dimension reduction (e.g., linear subspace models)~\cite{Aguiar:2010:SS,DBLP:journals/tog/HahnTCSCMDG14} or coarse-to-fine strategies are commonly applied~\cite{Wang:2010:EBW,Muller:2010:WM:1921427.1921441,Kavan:2011:PUC:2010324.1964988}.
DRAPE~\cite{DBLP:journals/tog/GuanRHWB12} automates the process of learning linear subspaces from simulated data and applying them to different subjects. The model factors out body shape and pose to produce a global cloth shape and then applies the wrinkles of seen garments. However, deformations are applied per triangle as in \cite{Sumner:2004:DTT:1015706.1015736}, which is not optimal for online applications. 
Additionally, for all these methods, simulated data is tedious to generate, and accuracy and realism are limited.

\subsubsection{Learning methods.}
Previously mentioned methods focus on efficients simulation and representation of previously seen data. Going a step further, several methods have attempted to generalize this knowledge to unseen cases.
\cite{DBLP:conf/cvpr/TungM13} learns bags of dynamical systems to represent and recognize repeating patterns in wrinkle deformations. 
In DeepGarment \cite{CGF:CGF13125} the global shape and low frequency details are reconstructed from a single segmented image using a CNN but no retargeting is possible.

Only sparse work has been done on learning to add realistic details to 3D surfaces with neural networks but several methods to enrich facial scans with texture exist
\cite{DBLP:conf/cvpr/SaitoWHNL17,DBLP:conf/iccv/OlszewskiLYZYHX17}.
In particular, Generative Adversarial Networks (GANs) \cite{DBLP:conf/nips/GoodfellowPMXWOCB14} are suitable to enhance low dimensional information with details. In \cite{DBLP:conf/iccv/LassnerPG17} it is used create realistic images of clothed people given a (possibly random) pose.

Outside of clothing, SR-GAN \cite{DBLP:conf/cvpr/LedigTHCCAATTWS17} solves the super-resolution problem of recovering photo-realistic textures from heavily downsampled images on public benchmark. The task has similarities to ours in generating high frequency details for coarse inputs but we use a content loss motivated by perceptual similarity instead of similarity in pixel space. \cite{smoke-simulation} uses a data-driven approach with a CNN to simulate highly detailed smoke flows.
Instead, pix2pix \cite{DBLP:conf/cvpr/IsolaZZE17} proposes a conditional GAN that creates realistic images from sketches or annotated regions or vice versa.
This design suits our problem better as we aim at learning and transferring underlying image structure.

In order to represent the highest possible level of detail at runtime,
we propose to revisit the traditional rendering pipeline of 3D engine with computer vision.
Our contributions take advantage of the normal mapping technique~\cite{DBLP:conf/siggraph/KrishnamurthyL96,DBLP:conf/siggraph/CohenOM98,DBLP:conf/visualization/CignoniMSR98}.
Note that displacement maps have been used to create wrinkle maps using texture information~\cite{Bee11,Fyffe:2017:MSC:3128975.3129003}. However, while results are visually good on faces, they still require high resolution mesh, and no temporal consistency is guaranteed across time. (Also, faces are arguably less difficult to track than clothing which are prone to occlusions and more loose.)

In this work, we claim the first entirely data-driven method that uses a deep neural network on normal maps to leverage 3D geometry of clothing.
  
\section{Deformation Subspace Model}\label{sec:subspace}
We model cloth deformations by learning a linear subspace model that factors out body pose and shape, as in \cite{DBLP:journals/tog/GuanRHWB12}.
However, our model is learned from real data, and deformations are applied per vertex for speed and flexibility regarding graphics pipelines \cite{DBLP:journals/tog/LoperM0PB15}.
Our strategy ensures deformations are represented compactly and with high realism.
First, we compute robust template-based non-rigid registrations from a 4D scan sequence (Sect.~\ref{subsec:data}), then a clothing deformation statistical model is derived (Sect.~\ref{subsec:statisticalmodel}) and finally, a regression model is learned for pose retargeting (Sect.~\ref{subsec:regression}).

\subsection{Data preparation}\label{subsec:data}

\paragraph{Data capture.} For each type of clothing, we capture 4D scan sequences at 60 fps (e.g., 10.8k frames for 3 min) of a subject in motion, and dressed in a full-body suit with one piece of clothing with colored boundaries on top. Each frame contists of a 3D surface mesh with around 200k vertices yielding very detailed folds on the surface but partially corrupted by holes and noise (see Fig.~\ref{fig:overview}a). 
This setup allows a simple color-based 3D clothing extraction. In addition, capturing only one garment prevents occlusions where clothing normally overlaps (e.g., waistbands) and clothings can be freely combined with each other.

\paragraph{Body tracking.} 3D body pose is estimated at each frame using a method in the spirit of \cite{Taylor12}.
We define a skeleton with $j$ joints described by $p_j$ parameters representing transformation and bone length. Joint parameters are also adjusted to body shape, which is estimated using~\cite{DBLP:journals/tog/LoperM0PB15,DBLP:conf/cvpr/ZhangPBP17}.
Posed human body is obtained using a linear blend skinning function $S: \mathbb{R}^{3 \times v} \times \mathbb{R}^{p_j} \to \mathbb{R}^{3 \times v}$ that transforms (any subset of) $v$ vertices of a 3D deformable human template in normalized pose (e.g., T-pose) to a pose defined by $j$ skeleton joints. 

\begin{figure}[t]
    \includegraphics[width=.99\linewidth]{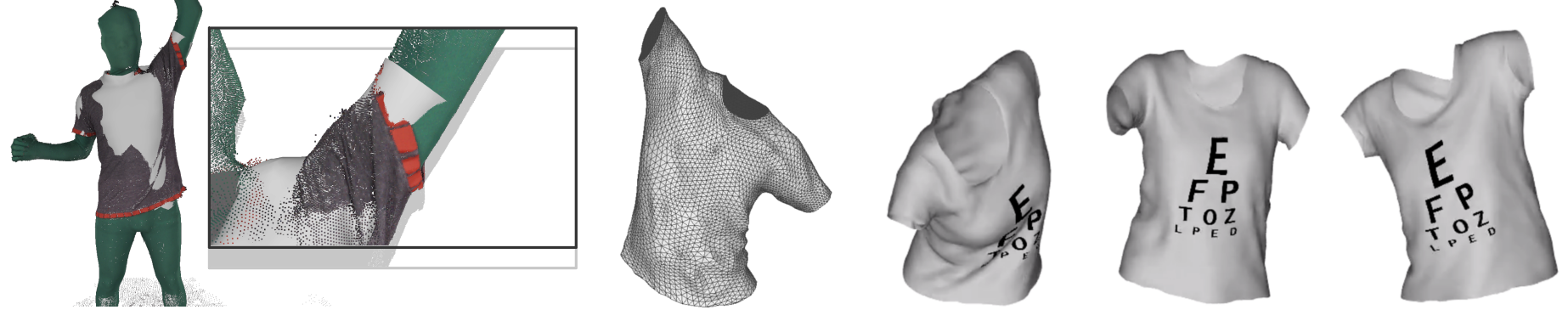}
    \caption{(Left) Before registration, a template is aligned to clothing scans by skinning. Boundaries are misplaced. (Right) Examples of registrations with different shirts on different people. Texture stability across sequence shows the method is robust to drift. } 
    \label{fig:notfitting}
\end{figure}

\paragraph{Registration.} We define a template of clothing $\bar{\mathcal{T}}$ by choosing a subset of the human template with consistent topology. $\bar{\mathcal{T}}$ should contain enough vertices to model deformations (e.g., 5k vertices for a T-shirt), as shown in Fig.~\ref{fig:notfitting}.
The clothing template is then registered to the 4D scan sequence using a variant of non-rigid ICP based on grid deformation~\cite{Li:2009:RSG:1618452.1618521,Guo2015RobustNM}.
The following objective function $\mathcal{E}_{reg}$, which aims at optimizing affine transformations of grid nodes, is iteratively minimized using Gauss-Newton method:
\begin{align}
\mathcal{E}_{reg} = \mathcal{E}_{data} + \omega_r\cdot\mathcal{E}_{rigid} + \omega_s\cdot\mathcal{E}_{smooth} + \omega_b\cdot\mathcal{E}_{bound},
\end{align}
where the data term $\mathcal{E}_{data}$ aligns template vertices with their nearest neighbors on the target scans, $\mathcal{E}_{rigid}$ encourages each triangle deformation to be as rigid as possible, and $\mathcal{E}_{smooth}$ penalizes inconsistent deformation of neighboring triangles.
In addition, we introduce the energy term $\mathcal{E}_{bound}$ to ensure alignment of boundary vertices,
which is unlikely to occur otherwise (see below for details).
We set $\omega_r=500$, $\omega_s=500$, and $\omega_b=10$ by experiments.
One template registration takes around 15s (using CPU only).

\begin{figure}[t]
\begin{center}
	\begin{overpic}
		[trim={1.5cm 2.7cm 0.9cm 2.9cm},clip,width=.4\linewidth]{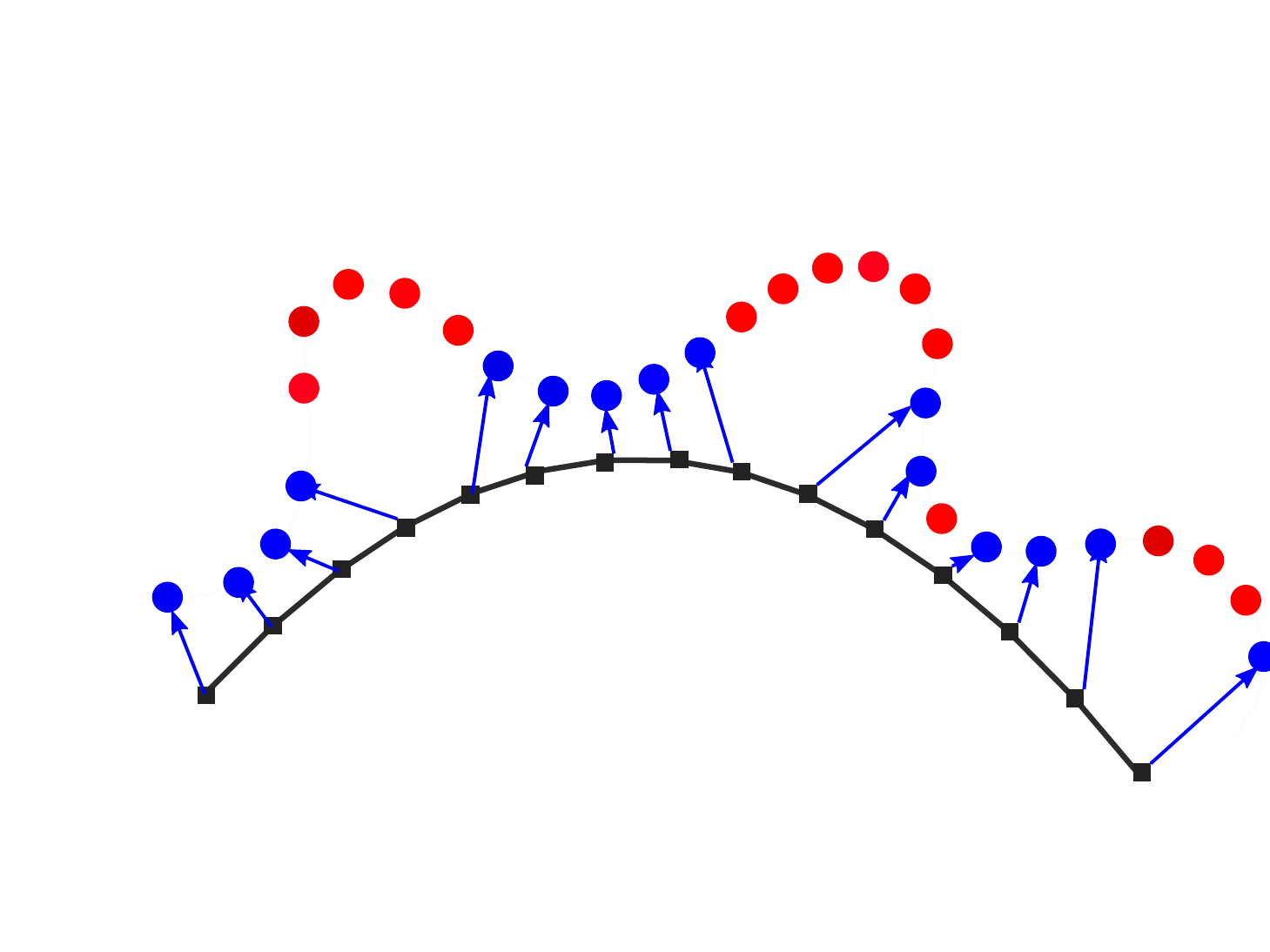}
    	\put(11,1){$\mathcal{B}_\mathcal{T}$}
        \put(79,33){$\mathcal{B}_\mathcal{S}$}
    \end{overpic}
    \quad\quad
    \begin{overpic}
    	[trim={1.5cm 2.7cm 0.9cm 2.9cm},clip,width=.4\linewidth]{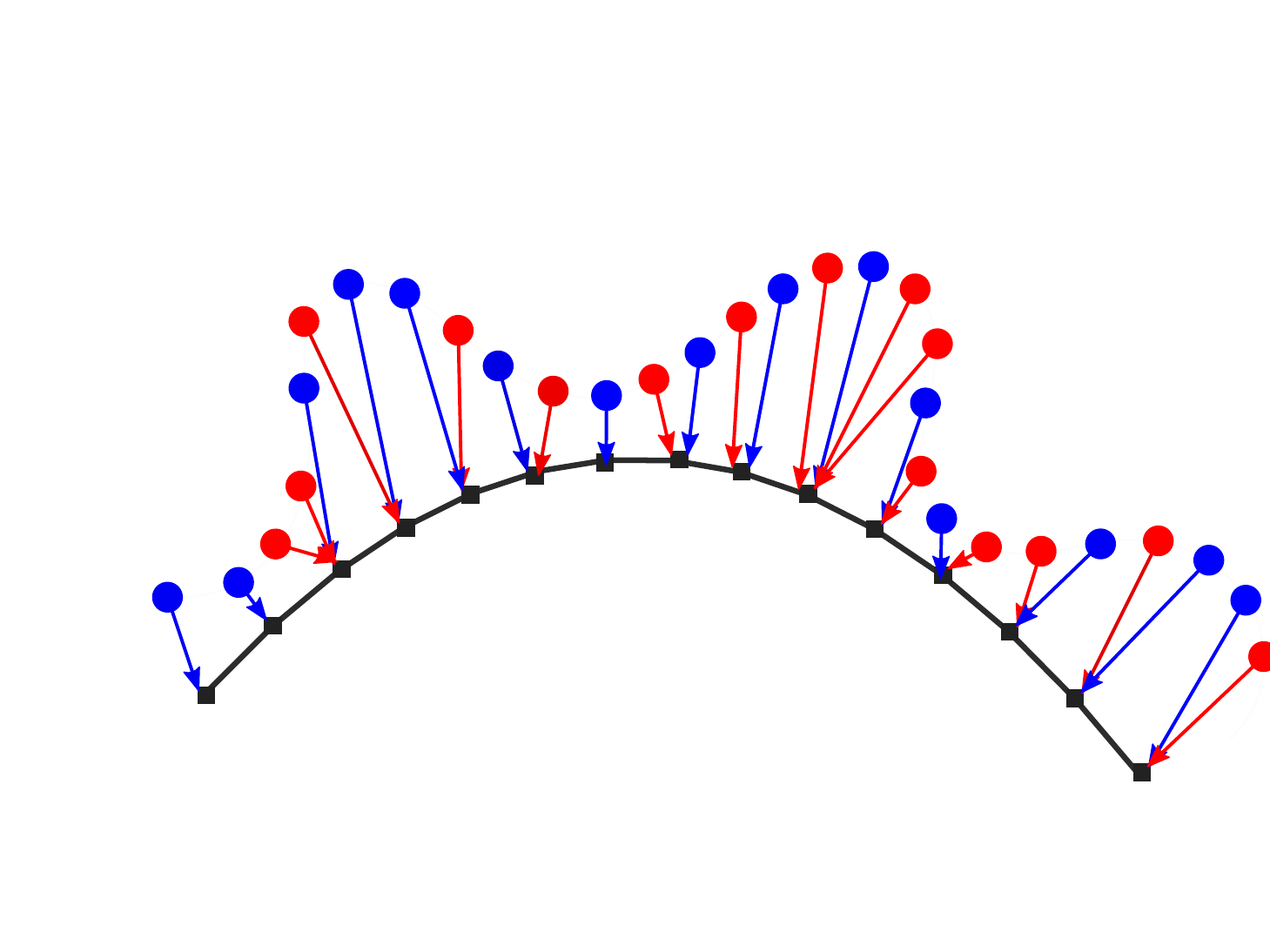}
        \put(11,1){$\mathcal{B}_\mathcal{T}$}
        \put(79,33){$\mathcal{B}_\mathcal{S}$}
    \end{overpic}
\end{center}
\caption{Strategies for Boundary Alignment. Template boundary points $\mathcal{B_T}$ are denoted in black, scan boundary points $\mathcal{B_S}$ (with significant more points than the template) are in red and blue. Points that are paired with a template point are in blue. (Left) Pairing to the closest neighbor of the template points leads to ignorance of distant details. (Right) Each template point in $\mathcal{B_T}$ is paired with the furthest point (marked in blue) in a set containing its closest points in $\mathcal{B_S}$.}
    \label{fig:manualpairs} 
\end{figure}

\paragraph{Boundary Alignment.}\label{subsec:boundarypairs} During data capture the boundaries of the clothing are marked in a distinguishable color and corresponding points are assigned to the set $\mathcal{B_S}$.
We call the set of boundary points on the template $\mathcal{B_T}$.
Matching point pairs in $\mathcal{B_S} \times \mathcal{B_T}$ should be distributed equally among the scan and template, and ideally capture all details in the folds. As this is not the case if each point in $\mathcal{B_T}$ is simply paired with the closest scan boundary point (see Fig.~\ref{fig:manualpairs}), we select instead a match $s_t \in \mathcal{B_S}$ for each point $t \in \mathcal{B_T}$ via the following formula:
\begin{align}
	s_t = \max_{s \in \mathcal{C}}\Vert t - s \Vert \quad \text{with } \quad \mathcal{C} = \left\{ s' \in \mathcal{B_S} \mid \arg \min_{t' \in \mathcal{B_T}} \Vert s' - t' \Vert = t \right\}.
\end{align}
Notice that $\mathcal{C}$ might be empty. This ensures consistency along the boundary and better captures high frequency details (which are potentially further away).

\subsection{Statistical model}\label{subsec:statisticalmodel}

The statistical model is computed using linear subspace decomposition by PCA~\cite{DBLP:journals/tog/LoperM0PB15}.
Poses $\{\mathcal{\theta}_1,...,\mathcal{\theta}_n\}$ of all $n$ registered meshes $\{\mathcal{R}_1,...,\mathcal{R}_n\}$ are factored out from the model by pose-normalization using inverse skinning:
$S^{-1}(\mathcal{R}_i, \mathcal{\theta}_i) = \bar{\mathcal{R}}_i \in \mathbb{R}^{3 \times v}$. In what remains, meshes in normalized pose are marked with a bar.
Each registration $\bar{\mathcal{R}}_i$ can be represented by a mean shape
$\bar{\mathcal{M}}$ and vertex offsets $o_i$, such that $\bar{\mathcal{R}}_i = \bar{\mathcal{M}} + o_i$,
where the mean shape $\bar{\mathcal{M}} \in \mathbb{R}^{3 \times v}$ is obtained by averaging vertex positions:	$\bar{\mathcal{M}} = \sum_{i = 1}^n \frac{\bar{\mathcal{R}}_i}{n}$.
The $n$ principal directions of the matrix $O = [ o_1\ \dotsc \ o_n ]$ are obtained by singular value decomposition: $O = U \Sigma V^\top$.
Ordered by the largest singular values, the corresponding singular vectors contain information about the most dominant deformations.

Finally, each $\mathcal{R}_i$ can be compactly represented by $k \leq n$ parameters $\{\lambda_1^i,...,\lambda_k^i\} \in \mathbb{R}^k$ (instead of its $3 \times v$ vertex coordinates), with the linear blend shape function $B$, given a pose $\mathcal{\theta}_i$:
\begin{equation}
B(\{\lambda_1^i,...,\lambda_k^i\}, \mathcal{\theta}_i) = S\left( \bar{\mathcal{M}} + \sum_{l=0}^k \lambda_l^i \cdot V_l, \mathcal{\theta}_i\right) \approx \mathcal{R}_i \in \mathbb{R}^{3 \times v}, \label{eq:blendshape}
\end{equation}
where $V_l$ is the $l$-th singular vector. For a given registration, $\lambda_l^i = V_l^\top \bar{\mathcal{R}}_i$ holds. In practice, choosing $k = 500$ is sufficient to represent all registrations with a negligible error (less than 5 mm).

\subsection{Pose-to-shape prediction}\label{subsec:regression}
We now learn a predictive model $f$, that takes as inputs $j$ joint poses, and outputs a set of $k$ shape parameters $\Lambda$.
This allows powerful applications where deformations are induced by pose.
To take into account deformation dynamics that occur during human motion, the model is also trained with pose velocity, acceleration, and shape parameter history. These inputs are concatenated in the control vector $\Theta$, and $f$ can be obtained using autoregressive models~\cite{Aguiar:2010:SS,Dyna:SIGGRAPH:2015,DBLP:journals/tog/LoperM0PB15}.

In our experiments with clothing, we solved for $f$ in a straightforward way by linear regression: $F = \Lambda \cdot \Theta^\dagger$,
where $F$ is the matrix representation of $f$, and $\dagger$ indicates the Moore-Penrose inverse.
While this allows for (limited) pose retargeting, we observed loss in reconstruction details. One reason is that under motion, the same pose can give rise to various configurations of folds depending on the direction of movement, speed and previous fold configurations.

To obtain non-linear mapping, we consider the components of $\Theta$ and $\Lambda$ as multivariate time series, and train a deep multi-layer recurrent neural network (RNN)~\cite{DBLP:conf/nips/SutskeverVL14}. A sequence-to-sequence encoder-decoder architecture with Long Short-term Memory (LSTM) units is well suited as it allows continuous predictions, while being easier to train than RNNs and outperforming shallow LSTMs.
We compose $\Theta$ with $j$ joint parameter poses, plus velocity and acceleration of the joint root. MSE compared to linear regression are reported in Sect.~\ref{subsec:retargetting}.

\begin{figure}[t]
	\includegraphics[width=.23\linewidth]{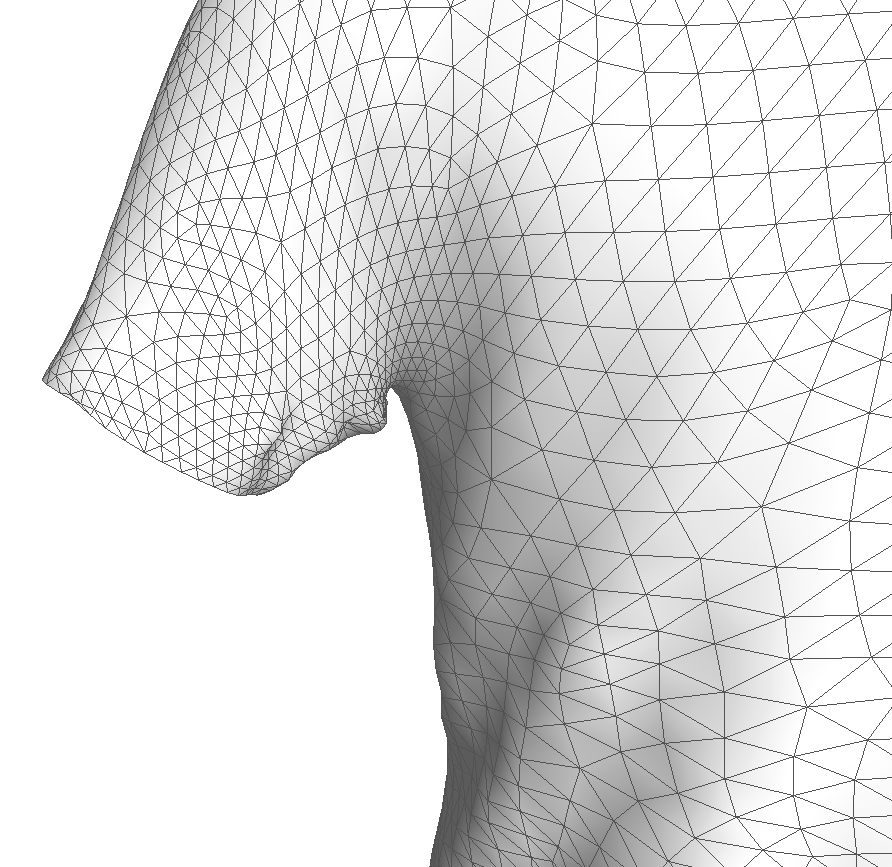}
    \includegraphics[width=.24\linewidth]{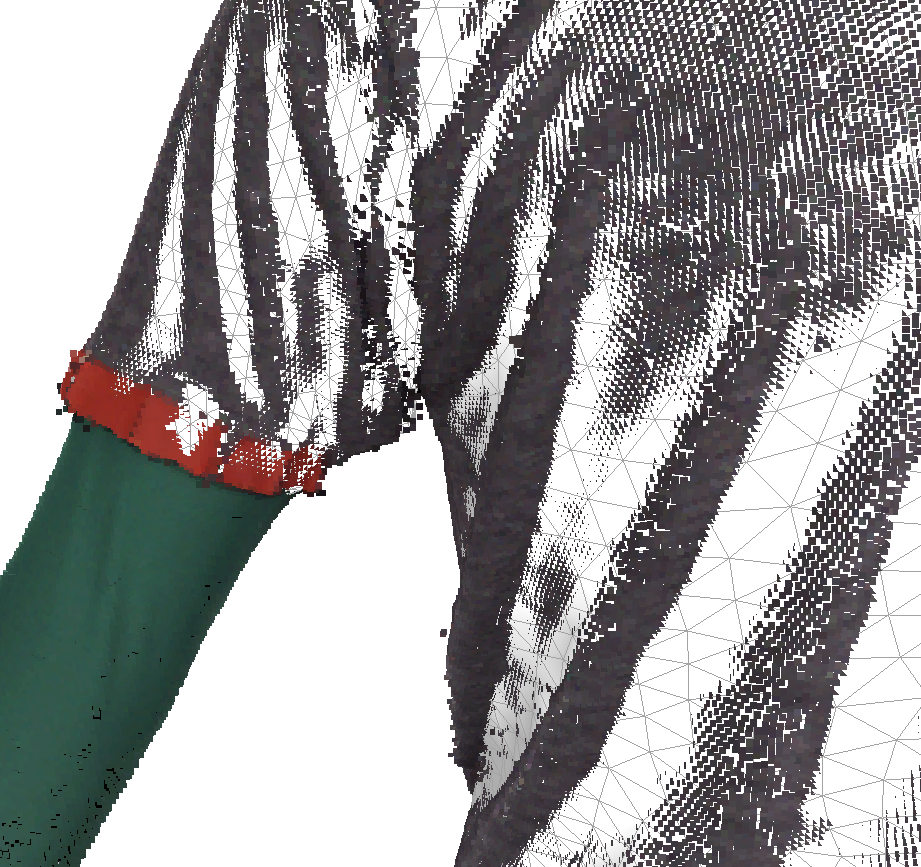}
    \includegraphics[width=.5\linewidth]{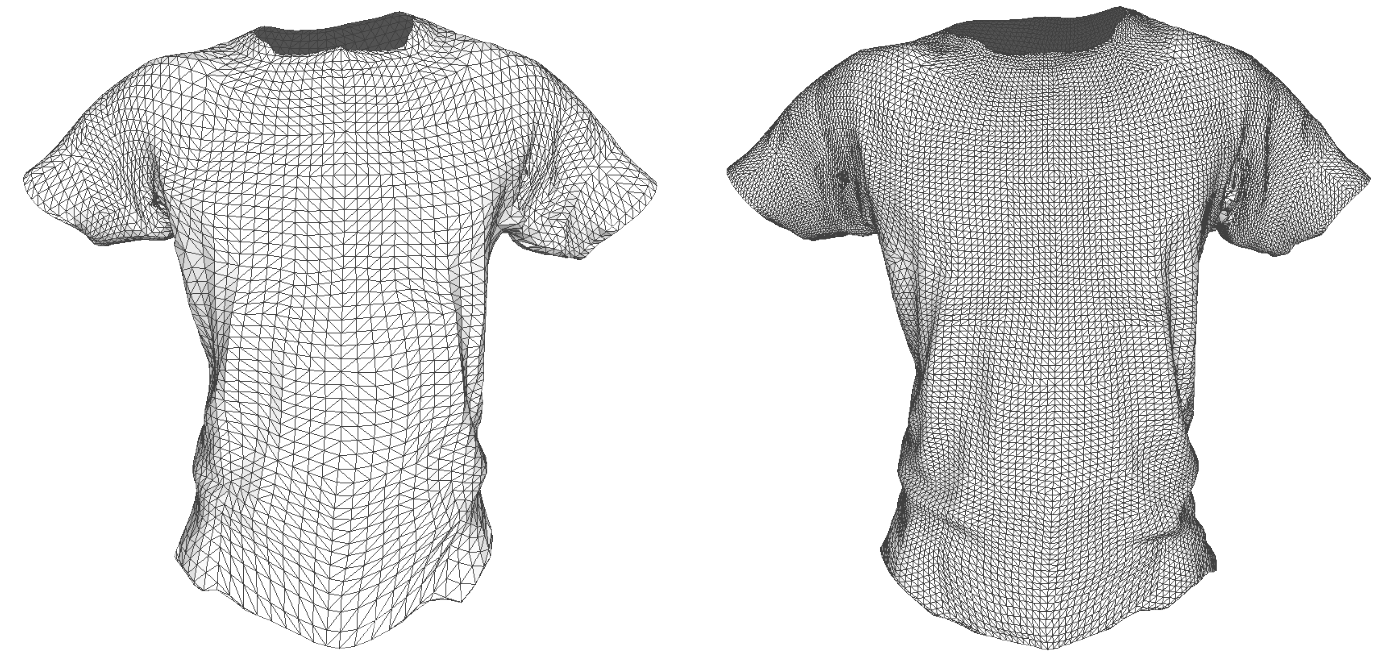}
    \caption{Limits of registration and subspace model. (Left) Global shape is well recovered, but many visible (high frequency) details are missing. (Right) Increasing the resolution of the template mesh is still not sufficient. Note that~\cite{ClothCap17} suffers from the same limitations.}
    \label{fig:bslimits}
\end{figure}

\section{Fine Wrinkle Generation}\label{sec:gan}
Our goal is to recover all observable geometric details.
As previously mentioned, template-based methods~\cite{ClothCap17} and subspace-based methods~\cite{DBLP:journals/tog/HahnTCSCMDG14,DBLP:journals/tog/GuanRHWB12} cannot recover every detail such as fine cloth wrinkles due to resolution and data scaling limitations, as illustrated in Fig.~\ref{fig:bslimits}.

\begin{figure}[t]
	\includegraphics[trim={0cm, 0.4cm, 7cm, 0cm},clip,width=.55\linewidth]{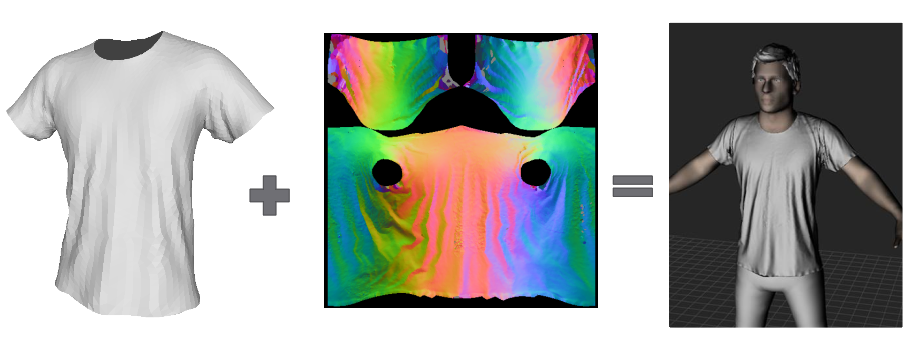}
    \includegraphics[trim={1cm, 0.7cm, 0.8cm, 0.5cm},clip,width=.23\linewidth]{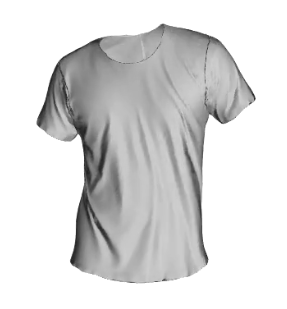}
    \includegraphics[trim={2cm, 0cm, 0.5cm, 0cm}, clip, width=.19\linewidth]{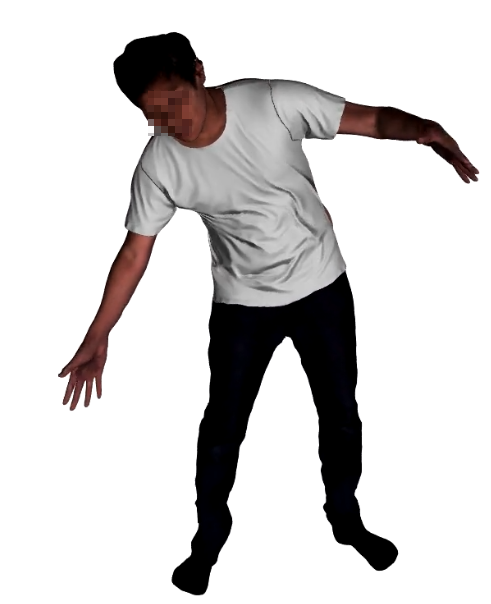}
    \caption{All visible details from an accurate 3D scan are generated in our normal map for incredible realism. Here, a virtual shirt is seamlessly added on top of an animated virtual human (e.g., scanned subject).}
    \label{fig:normalmaps}
\end{figure}

Assuming the finest details are captured at sensor image pixel resolution, and are reconstructed in 3D (e.g., using a 4D scanner as in~\cite{dfaust:CVPR:2017,ClothCap17}), all existing geometric details can then be encoded in a normal map of the 3D scan surface at lower resolution (see Fig~\ref{fig:normalmaps}).
To automatically add fine details \emph{on the fly} to reconstructed clothing, we propose to leverage normal maps using a generative adversarial network~\cite{DBLP:conf/nips/GoodfellowPMXWOCB14}. See Figure~\ref{fig:cgan} for the architecture.
In particular, our network induces temporal consistency on the normal maps to increase realism in animation applications.

\subsection{Data preparation}

\begin{figure}[t]
	\includegraphics[width=.32\linewidth]{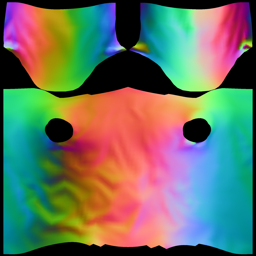}
    \includegraphics[width=.32\linewidth]{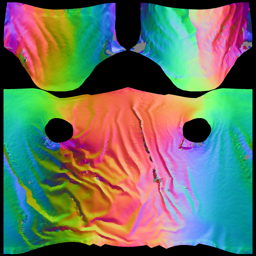}
    \includegraphics[trim={1.5cm 2.7cm 0.9cm 1.3cm},clip,width=.16\linewidth]{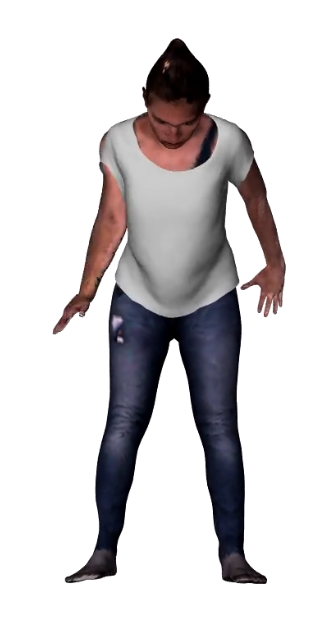}
    \includegraphics[trim={1.5cm 2.7cm 0.9cm 1.3cm},clip,width=.16\linewidth]{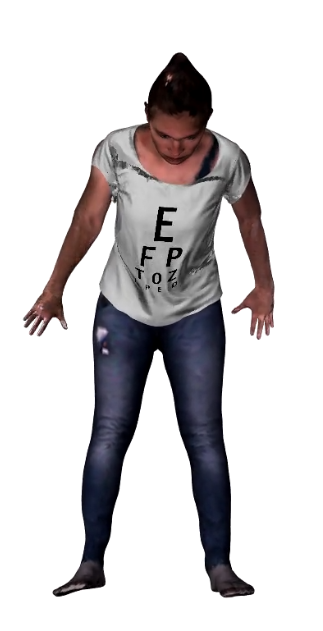}
    \caption{Examples of our dataset. (Left) Low resolution input normal map, (Middle) High resolution target normal map from scan. Details, and noise, visible on the scan are reproduced in the image. Gray areas indicate no normal information was available on the scan. (Right) T-Shirt on a human model rendered without and with normal map. }
    \label{fig:dataset}
\end{figure}

We take as inputs a 4D scan sequence, and a sequence of corresponding reconstructed garments. The latter can be either obtained by registration, reconstruction using blend shape or regression, as detailed in Sect.~\ref{sec:subspace}.
Clothing template meshes $\bar{\mathcal{T}}$ are equipped with UV maps which are used to project any pixel from an image to a point on a mesh surface,
hence assigning a property encoded in a pixel to each point. Therefore, normal coordinates can be normalized and stored as pixel colors in normal maps.
Our training dataset then consists of pairs of normal maps (see Fig.~\ref{fig:dataset}):
\emph{low resolution} (LR) normal maps obtained by blend shape reconstruction,
and \emph{high resolution} (HR) normal maps obtained from the scans.
For LR normal maps, the normal at surface point (lying in a face) is linearly interpolated from vertex normals.
For HR normal maps, per-pixel normals are obtained by projection of the high resolution observations (i.e., 4D scan) onto triangles of the corresponding low resolution reconstruction, and then the normal information is transferred using the UV map of $\bar{\mathcal{T}}$.
Note that normal maps cannot be directly calculated from scans because neither is the exact area of the garment defined, nor are they equipped with UV map.
Also, our normals are represented in global coordinates, as opposed to tangent space coordinates as is standard for normal maps.
The reason is that LR normal maps contain no additional information to the geometry and are therefore constant in tangent space.
This makes them unfitting for conditioning our adversarial neural network.

\subsection{Network architecture}
Due to the nature of our problem it is natural to explore network architectures designed to enhance images (i.e., super-resolution applications). From our experiments, we observed that models trained on natural images, including those containing a perceptual loss term fail (e.g., SR-GAN~\cite{DBLP:conf/cvpr/LedigTHCCAATTWS17}).
On the other hand, cloth deformations exhibit smooth patterns (wrinkles, creases, folds) that deform continuously in time. In addition, at a finer level, materials and fabric texture also contain high frequency details.

\begin{figure}[t]
	\input{images/cGAN.tikz}
    \caption{cGAN for realistic HR normal map generation from LR normal maps as input. Layer sizes are squared. Skip connections (red) in U-Net preserve underlying image structure across network layers. PatchGAN enforces wrinkle pattern consistency.}
    \label{fig:cgan}
\end{figure}
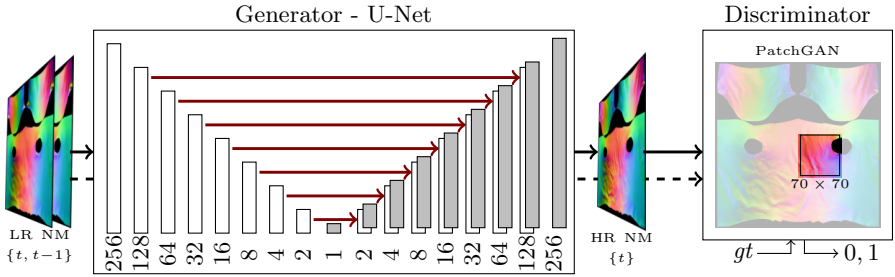

Our proposed network is based on a conditional Generative Adversarial Network (cGAN) inspired from image transfer~\cite{DBLP:conf/cvpr/IsolaZZE17}. We also use a convolution-batchnorm-ReLu structure \cite{Ioffe:2015:BNA:3045118.3045167} and a U-Net in the generative network since we want latent information to be transfered across the network layers and the overall structure of the image to be preserved. This happens thanks to the skip connections.
The discriminator only penalizes structure at the scale of patches, and works as a texture loss. 
Our network is conditioned by low-resolution normal map images (size: $256 \times 256$) which will be enhanced with fine details learned from our real data normal maps. See Fig.~\ref{fig:cgan} for the complete architecture.

\begin{figure}[t]
	\includegraphics[width=.23\linewidth]{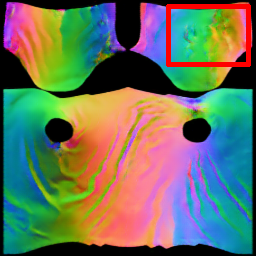}
    \includegraphics[width=.23\linewidth]{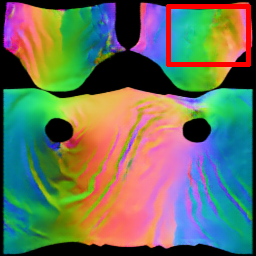}
    \quad
    \includegraphics[width=.23\linewidth]{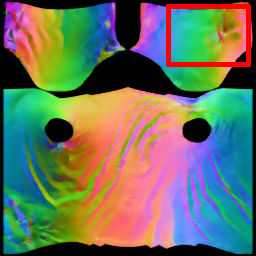}
    \includegraphics[width=.23\linewidth]{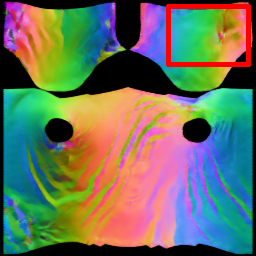}
    \caption{Examples trained on only 2000 training samples reinforce the effect of the additional loss $\mathcal{L}_{temp}$. The pairs show twice the same consecutive frames: (left) \emph{without} temporal consistency term geometric noise appears or disappears instantly, and (right) \emph{with} temporal consistency term preserves geometric continuity.
    } 
    \label{fig:temporalterm}
\end{figure}

Temporal consistency is achieved by extending the $L1$ network loss term. For compelling animations, it is not only important that each frame looks realistic, but also no sudden jumps in the rendering should occur.
To ensure smooth transition between consecutively generated images across time, we introduce an additional loss $\mathcal{L}_{loss}$ to the GAN objective that penalizes discrepancies between generated images at $t$ and expected images (from training dataset) at $t-1$:
\begin{align}
\mathcal{L}_{loss} =\ \underbrace{\Vert \mathcal{I}_{gen}^t - \mathcal{I}_{gt}^t \Vert_1}_{\mathcal{L}_{data}} + \underbrace{\mid \sum_{i,j} (\mathcal{I}_{gen}^t - \mathcal{I}_{gt}^{t-1})_{i,j} \mid}_{\mathcal{L}_{temp}},
\label{eq:temporal}
\end{align}
where $\mathcal{L}_{data}$ helps to generate images near to ground truth in an $L_1$ sense (for less blurring).
The temporal consistency term $\mathcal{L}_{temp}$ is meant to capture global fold movements over the surface. If something appears somewhere, most of the time, it should have disappeared close-by and vice versa. Our term does not take spatial proximity into account though. We also tried temporal consistency based on the $L_1$- and $L_2$-norm, and report the results in Table~\ref{tab:errors}. See Fig.~\ref{fig:temporalterm} for a comparison of results with and without the temporal consistency term.

\section{Experiments}\label{sec:experiments}

This section evaluates the results of our reconstruction.
4D scan sequences were captured using a temporal-3dMD system (4D)~\cite{3dmd}.
Sequences are captured at 60fps. Each frame consists of a colored mesh with 200K vertices.
Here, we show results on two different shirts (for female and male).
We trained the cGAN network on a dataset of 9213 consecutive frames. The first 8000 images compose the training data set, the next 1000 images the test data set and the remaining 213 images the validation set. Test and validation sets contain poses and movements not seen in the training set.
The U-Net auto-encoder is constructed with $2\times 8$ layers, and 64 filters in each of the first convolutional layers. The discriminator uses patches of size $70 \times 70$.
$\mathcal{L}_{data}$ weight is set to 100, $\mathcal{L}_{temp}$ weight is 50, while GAN weight is 1.
The images have a resolution of $256 \times 256$, although our early experiments also showed promising results on $512 \times 512$.
 
\subsection{Comparison of approaches}

\begin{figure}[h]
\begin{center}
\includegraphics[width=0.98\linewidth]{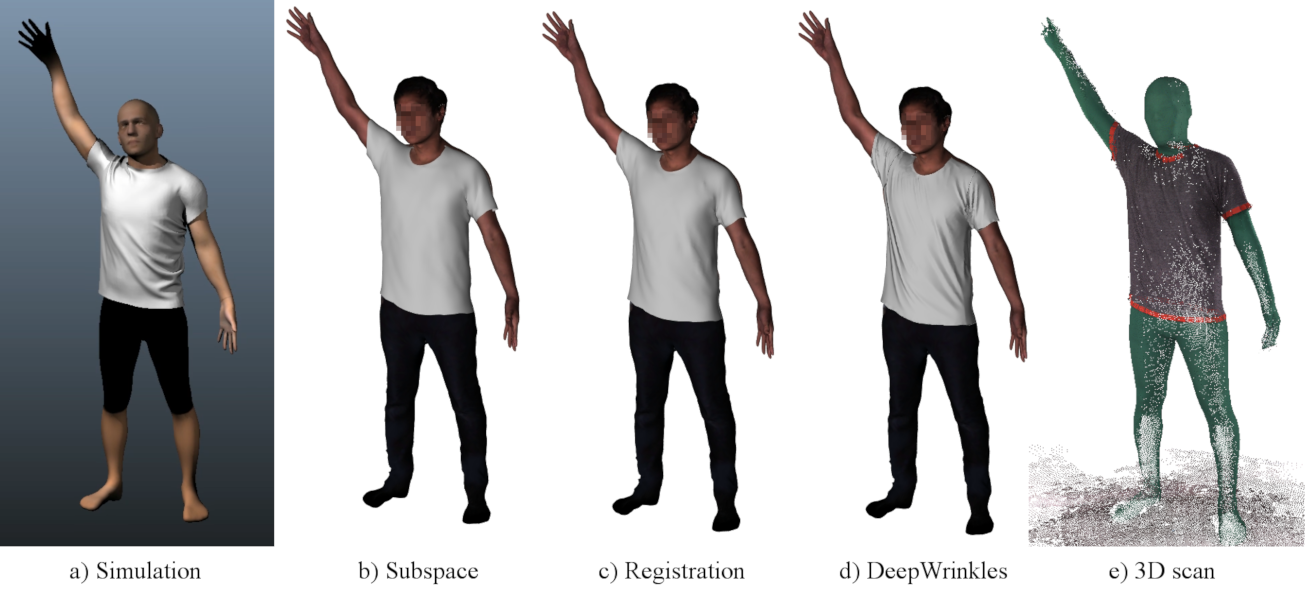}
    \caption{Comparison of approaches. a) Physics-based simulation \cite{MarvelousDesigner}, b) Subspace (50 coefficients)~\cite{DBLP:journals/tog/GuanRHWB12}, c) Registration~\cite{ClothCap17}, d) DeepWrinkles (ours), e) 3D scan (ground truth).} 
    \label{fig:simulation}
\end{center}
\end{figure}

We compare our results to different approaches (see Fig. \ref{fig:simulation}).
A physics-based simulation done by a 3D artist using MarvelousDesigner~\cite{MarvelousDesigner} returns a mesh imitating similar material properties as our scan and with a comparable amount of folds but containing $53,518$ vertices (i.e., an order of magnitude more). A linear subspace reconstruction with $50$ coefficients derived from the registrations (Sect. \ref{sec:subspace}) produces a mostly flat surface, while the registration itself shows smooth approximations of the major folds in the scan.
Our method, DeepWrinkles, adds all high frequency details seen in the scan to the reconstructed surface. These three methods use a mesh with $5,048$ vertices. DeepWrinkles is shown with a $256 \times 256$ normal map image.

\begin{table}[h]
\centering
    \begin{tabular}{l || l | l | l | l || l | l | l | l  }
    	 & $L1$ temp \quad & $L2$ temp \quad & Eq. 
         \ref{eq:temporal} \quad & no temp \quad & Registr. \quad & BS 500 \quad & BS 200\quad  & Regre. \quad \\ \hline
    	Data & 4.63 & 5.16 & 3.72 & 5.06 & 3.72 & 6.86 & 6.40 & 7.6  \\
        Temporal\quad & 5.11 & 4.2 & 4.2 & 5.52 & 4.2 & 7.05 & 6.52 & 7.65 
    \end{tabular}
    \caption{Comparison of pixel-wise error values of the neural network for different training types. Data and Temporal are as defined in Eq.~\ref{eq:temporal}. (Left) Different temporal consistency terms. $L1$ and $L2$ take the respective distance between the output and target at time $t-1$. (Right) Different reconstruction methods to produce the input normal map. Registr. refers to registration, BS to the blend shape with a certain number of basis functions and Regre. to regression. } \label{tab:errors}
\end{table}

\begin{figure}[h]
\begin{overpic}
    [width=.16\linewidth]{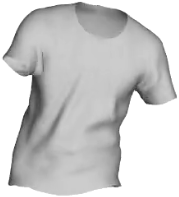}
    \put(-8,-13){\tiny{\textls[-50]{a) No Normal Map}}}
\end{overpic}
\begin{overpic}
    [width=.16\linewidth]{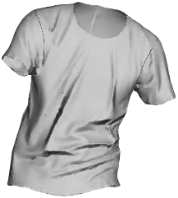}
    \put(-8,-13){\tiny{\textls[-50]{b) Ground-Truth}}}
\end{overpic}
\begin{overpic}
    [width=.16\linewidth]{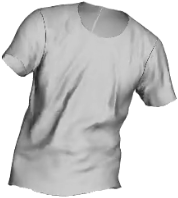}
    \put(-20,-13){\tiny{\textls[-50]{c) Reg. with temporal}}}
\end{overpic}
\begin{overpic}
    [width=.16\linewidth]{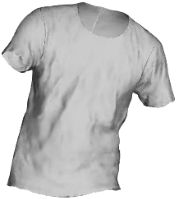}
    \put(-12,-13){\tiny{\textls[-50]{d) 200 with temporal}}}
\end{overpic}
\begin{overpic}
    [width=.16\linewidth]{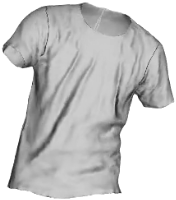} 
    \put(-8,-13){\tiny{\textls[-50]{e) 500 with temporal}}}
\end{overpic}
\begin{overpic}
    [width=.16\linewidth]{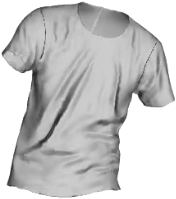}
    \put(-1,-13){\tiny{\textls[-50]{f) Reg. no temporal}}}
\end{overpic}
    \caption{Examples of different training results for high resolution normal maps. Left to right: Global shape, target normal map, learned from registration normal map with temporal consistency, learned from blend shape with 200 basis functions and temporal consistency, as previous but with 500 basis functions, learned from registration normal map without temporal consistency. The example pose is not seen in the training set. }
    \label{fig:preprocessing}
\end{figure}

\subsection{Importance of reconstruction details in input}

Our initial experiments showed promising results reconstructing details from the original registration normal maps. To show the efficacy of the method it is not only necessary to reconstruct details from registration, but also from blend shapes, and after regression. We replaced the input images in the training set by normal maps constructed from the blend shapes with 500, 200 and 100 basis functions and one set from the regression reconstruction. The goal is to determine the amount of detail that is necessary in the input to obtain realistic detailed wrinkles. 
Table~\ref{tab:errors} shows the error rates in each experiment. $500$ basis functions seem sufficient for a reasonable amount of detail in the result. Probably due to the fact that the reconstruction from regression is more noisy and bumpy, the neural network is not capable of reconstructing long defined folds and instead produces a lot of higher frequency wrinkles (see Fig.~\ref{fig:preprocessing}). This is an indicator that the structures of the inputs are only redefined by the net and important folds have to be visible in the input.

\subsection{Retargeting}\label{subsec:retargetting}

The final goal is to be able to scan one piece of clothing in one or several sequences and then transferring it on new persons with new movements on the go. 

\paragraph{Poses.}
We experimented with various combinations of control vectors $\Theta$, including pose, shape, joint root velocity and acceleration history. 
It turns out most formulations in the literature are difficult to train or unstable~\cite{Aguiar:2010:SS,Dyna:SIGGRAPH:2015,DBLP:journals/tog/LoperM0PB15}. We restrict the joint parameters to those directly related to each piece of clothing to reduce the dimensionality. In the case of shirts, this leaves the parameters related to the upper body.
In general, linear regression generalized best but smoothed out a lot of overall geometric details, even in the training set.
We evaluated on 9213 frames for 500 and 1000 blend shapes: $MSE_{500} = 2.902$ and $MSE_{1000} = 3.114$.

On the other hand, we trained an encoder-decoder with LSTM units (4 layers with dimension 256), using
inputs and outputs equally of length 3 (see Sect.~\ref{subsec:regression}).
We obtained promising results: $MSE_{rnn} = 1.892$.
Supplemental materials show visually good reconstructed sequences.

\paragraph{Shapes.}
In \ref{subsec:statisticalmodel} we represented clothing with folds as offsets of a mean shape. The same can be done with a human template for persons with different body shapes. Each person $\bar{\mathcal{P}_i}$ in normalized pose can be represented as an average template plus a vertex-wise offset $\bar{\mathcal{P}_i} = \bar{\mathcal{T'}} + o'_i$. Given the fact that the clothing mean shape $\bar{\mathcal{M}} = \bar{\mathcal{T'}}_{\mid \mathcal{M}} + o'_{\mid \mathcal{M}}$ contains a subset of vertices of the human template, it can be adjusted to any deformation of the template by taking  $\bar{\mathcal{M}}_{o'}= \bar{\mathcal{M}} + {o'_i}_{\mid \mathcal{M}}$. $\mid \mathcal{M}$ restricts vertices of the human template to those used for clothing. Then the mean in the blend shape can simply be replaced by $\bar{\mathcal{M}}_{o'}$.
Equation \ref{eq:blendshape} becomes:

\begin{equation}
	B(\{\lambda_1^i,...,\lambda_k^i\}, \mathcal{\theta}_i) = S\left( \bar{\mathcal{M}}
    _{o'} + \sum_{l=0}^k \lambda_l^i \cdot V_l, \mathcal{\theta}_i\right) \approx {\mathcal{P}_i}_{\mid \mathcal{M}},
\end{equation}

\begin{figure}[h]
    \includegraphics[width=.99\linewidth]{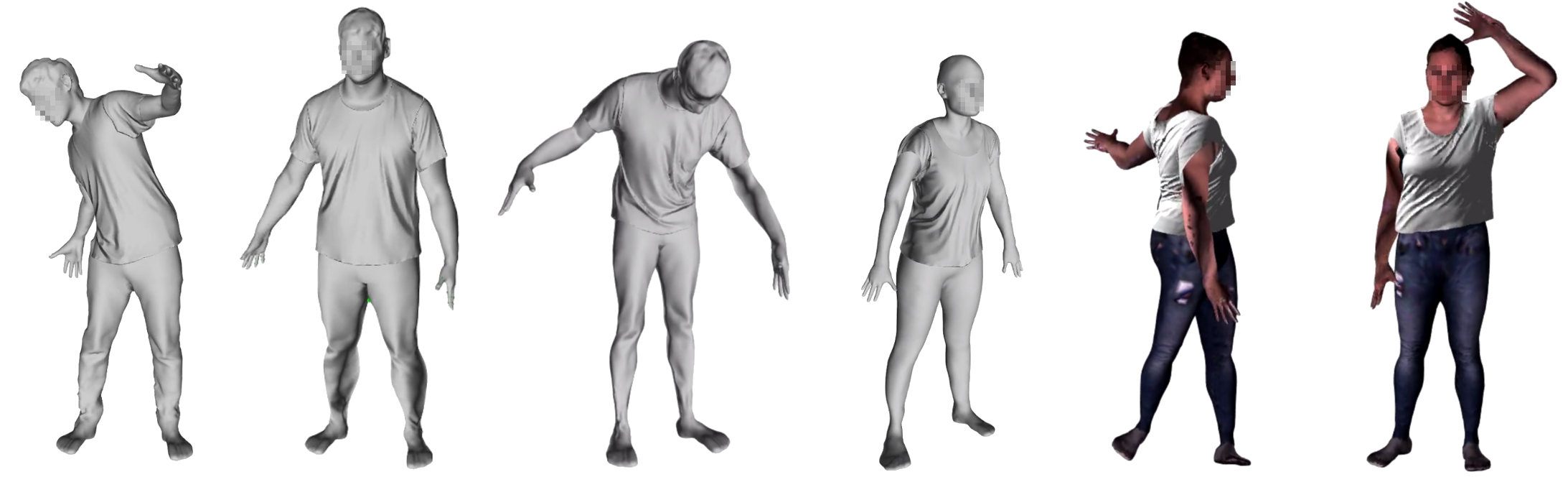}
    \caption{Body shape retargeting. The first and fourth entry are shirts on the original models, the following two are retargeted to new body shapes. }
    \label{fig:retargeting}
\end{figure}

Replacing the mean shape affects surface normals. Hence, it is necessary to use normal maps in tangent space at rendering time. This makes them applicable to any body shape (see Fig.~\ref{fig:retargeting}). 

\newpage 
\section{Conclusion}\label{sec:conclusion}

We present DeepWrinkles, a entirely data-driven framework to capture and reconstruct clothing in motion out from 4D scan sequences.
Our evaluations show that high frequency details can be added to low resolution normal maps
using a conditional adversarial neural network.
We introduce an additional temporal loss to the GAN objective that preserves geometric consistency across time, and show qualitative and quantitative evaluations on different datasets.
We also give details on how to create low resolution normal maps from registered data,
as it turns out registration fidelity is crucial for the cGAN training.
The two presented modules are complementary to achieve accurate and realistic rendering of global shape and details of clothing.
To the best of our knowledge, our methods exceeds the level of detail of the current state of the art in both physics-based simulation and data-driven approaches by far.
Additionally, the space requirement of a normal map is negligible in comparison to increasing the resolution of clothing mesh, which makes our pipeline suitable to standard 3D engines.

\paragraph{Limitations. }\label{subsec:limitations}
High resolution normal maps can have missing information in areas not seen by cameras, such as armpit areas. Hence, visually disruptive artifacts can occur although the clothing template can fix most of the issues (e.g., by doing a pass of smoothing).
At the moment pose retargeting works best when new poses are similar to ones included in the training dataset.
Although the neural network is able to generalize to some unseen poses, reconstructing the global shape from a new joint parameter sequence can be challenging. This should be fixed by scaling the dataset.

\paragraph{Future Work. }\label{subsec:futurework} 
Scanning setup can be extended to reconstruct all body parts with sufficient details without occlusions, and apply our method to more diverse types of clothing and accessories like coats, scarfs. Normal maps could also be used to add fine details like buttons which are hard to capture in 3D.\\

\noindent {\small{\textbf{Acknowledgements.}} We would like to thank the FRL teams for their support, and Vignesh Ganapathi-Subramanian for preliminary work on the subspace model.}

%
%
%
\bibliographystyle{splncs04}
\bibliography{egbib}

\end{document}

%% file: overview2.tikz
\tikzset{
>=stealth',
  diagramEl/.style={
    rectangle, 
    rounded corners, 
    draw=black, very thick,
    text width=7em, 
    minimum height=3em, 
    text centered},
diagramHead/.style={
    rectangle, 
    rounded corners, 
    draw=black, thick,
    text width=6em, 
    minimum height=1.5em, 
    text centered,
    rotate=90},
  line/.style={draw, thick, <-}
}
  
\begin{tikzpicture}
 	 \node [diagramHead] (learning)  {Learning};
     \node [diagramEl,violet, below right = -6em and 1em of learning] (scan)  {\scriptsize{4D Scan Sequence} \\ \includegraphics[trim={7cm 4cm 8cm 2.2cm},clip,height=6em]{images/scan215900.png}};
     \node [diagramEl,violet, below = .5em of scan] (template)  {\scriptsize{Cloth Template} \\ \includegraphics[trim={5cm 1cm 5cm 1cm},clip,height=4em]{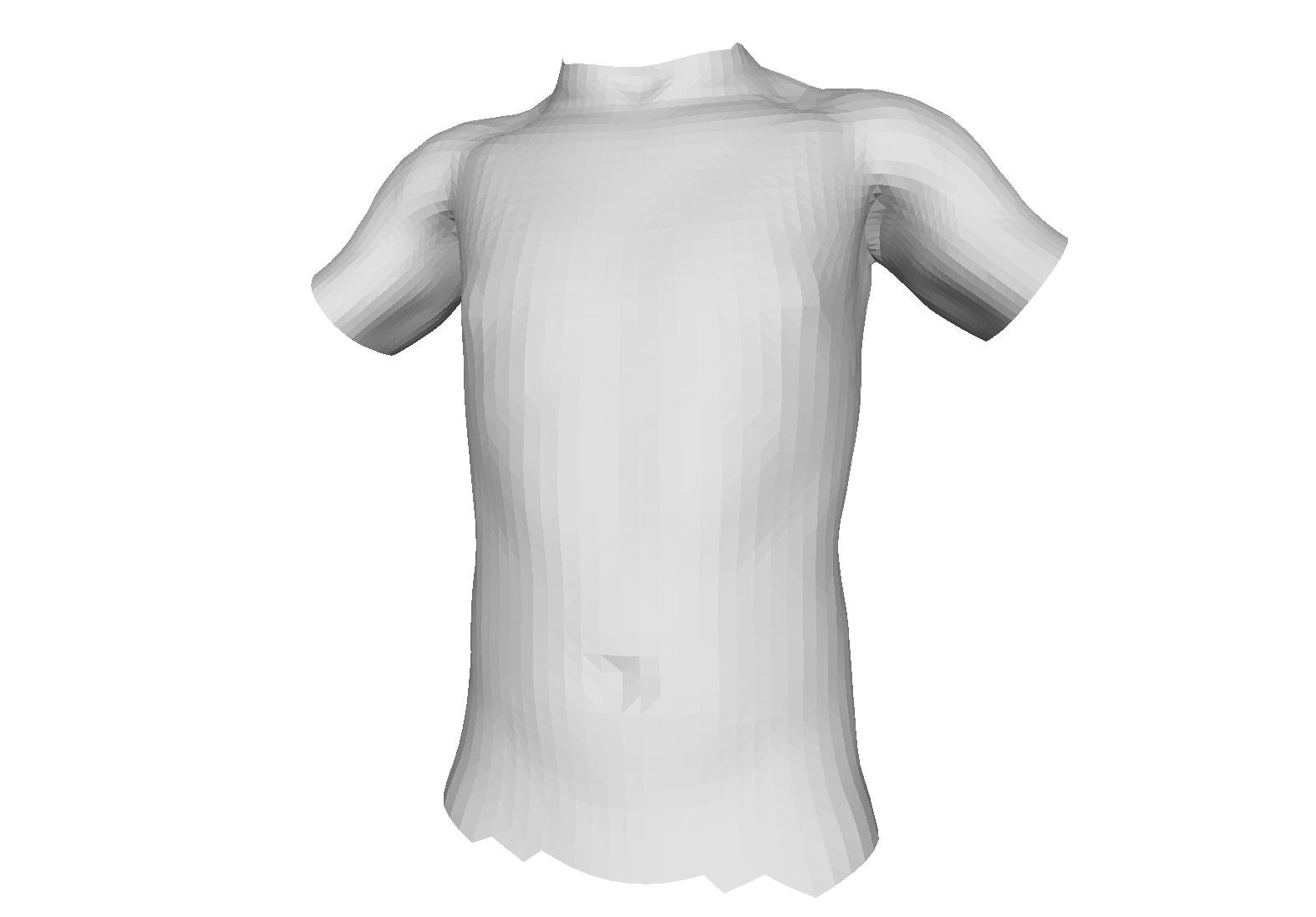}};
     \node [diagramEl, below = .5em of template] (tracking)  {\scriptsize{Body Shape \& Pose Estimation}}; 
     \node [diagramEl, below right = -6.5em and 1em of scan] (normalization)  {\scriptsize{Pose\\ Normalization} \\ \includegraphics[trim={10cm 3cm 10cm 1cm},clip,height=5em]{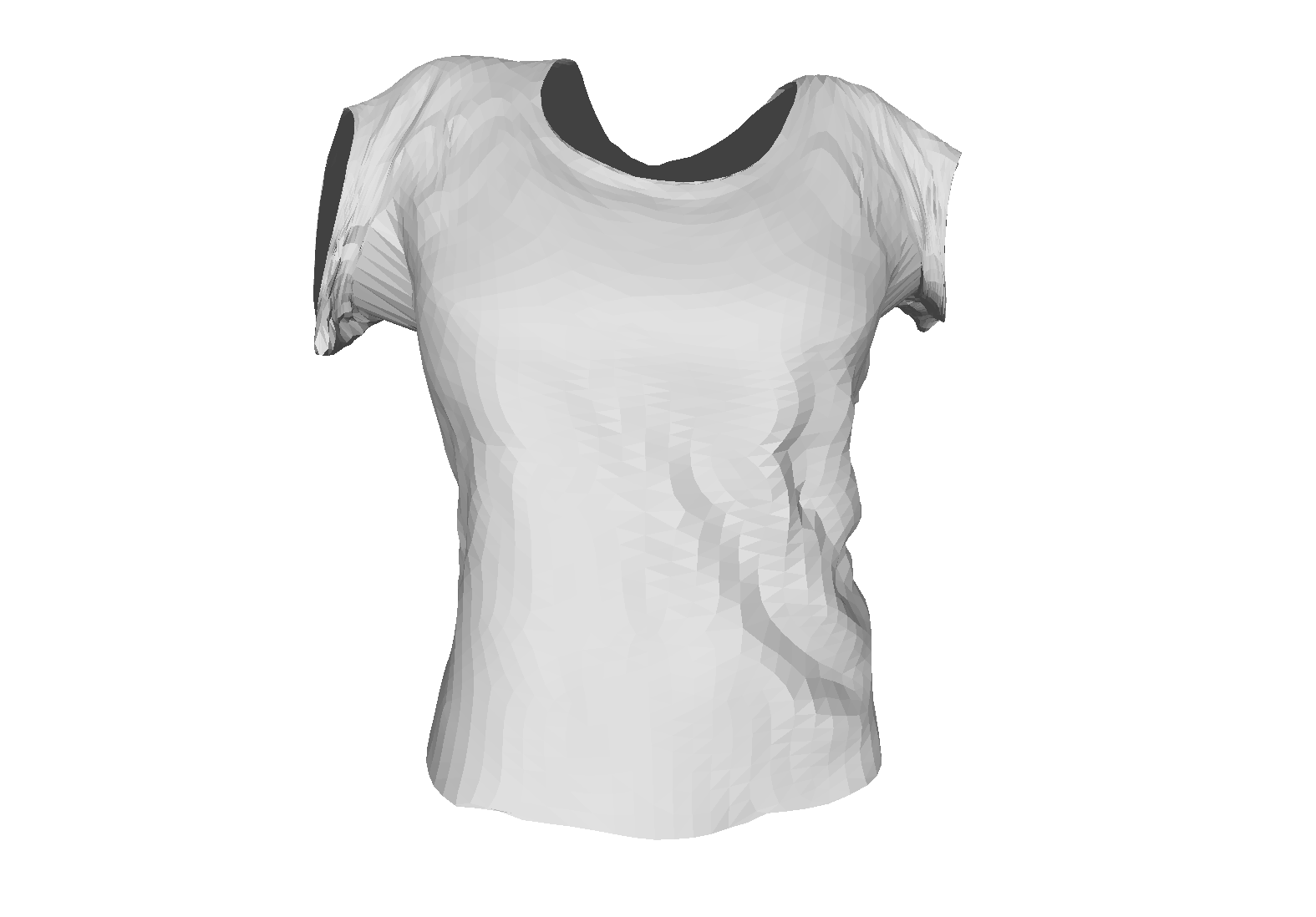}};
     \node [diagramEl, below = 1em of normalization] (registration)  {\scriptsize{Registration} \\ \includegraphics[trim={12cm 3cm 10cm 0cm},clip,height=5em]{images/registration03.png}};
     \node [diagramEl,teal, right = 1em of normalization] (blendshape)  {Statistical Model};
     \node [diagramEl,teal, below = 1em of blendshape] (regression)  {Blend shape Reconstruction};
     \node [diagramEl, below right = -5em and 1em of blendshape] (hr)  {\scriptsize{HR Normal Map} \\ \includegraphics[height=5em]{images/target.png}};
     \node [diagramEl, below = .5em of hr] (lr)  {\scriptsize{LR Normal Map} \\ \includegraphics[height=5em]{images/input.png}};
     \node [diagramEl,teal, below = 1em of lr] (cgan)  {cGAN};
     
     \draw[|-,-|,->, thick,] (registration.north) to (normalization.south);
     \draw[|-,-|,->, thick,] (blendshape.south) to (regression.north);
     \draw[|-,-|,->, thick,] (normalization.east) to (blendshape.west);
     \draw[|-,-|,->, thick,] (scan.north) |-(10,9.7em)-| (hr.north);
     \draw[|-,-|,->, thick,] (scan.west) -|(0.37,2em)|- (tracking.west);
     \draw[|-,-|,->, thick,] (scan.east) -|(3.2,-2em)|- (registration.west);
     \draw[|-,-|,->, thick,] (template.east) -|(3.2,-2em)|- (registration.west);
     \draw[|-,-|,->, thick,] (tracking.east) -|(3.2,-2em)|- (registration.west);
     \draw[|-,-|,->, thick,] ($(tracking.east)+(0,-0.3)$) -| (regression.south);
     \draw[|-,-|,->, thick,] (hr.east) -|($(hr.east)+(0.2,-2)$)|- (cgan.east);
     \draw[|-,-|,->, thick,] (regression.east) -|($(regression.east)+(0.1,1)$)|- (hr.west);
     \draw[|-,-|,->, thick,] (regression.east) -|($(regression.east)+(0.1,-0.1)$)|- (lr.west);
     \draw[|-,-|,->, thick,] (lr.south) to (cgan.north);
     
     \draw[gray,thick,solid] ($(scan.north west)+(-0.28,0.2)$) rectangle ($(cgan.south east)+(0.3,-0.11)$);
\end{tikzpicture}

\begin{tikzpicture}
	\node [diagramHead] (runtime)  {Runtime};
    \node [diagramEl, violet, below right = .5em and 1em of runtime] (estimation)  {\scriptsize{Body Shape \& Pose}};
    \node [diagramEl, teal, below = .5em of estimation] (statmodel)  {Statistical Model};
    \node [diagramEl, teal, right = 1em of estimation] (reconstruction)  {Blend shape Reconstruction};
    \node [diagramEl, below = 1em of reconstruction] (lrmap)  {\scriptsize{LR Normal Map}};
    \node [diagramEl, teal, right = 1em of reconstruction] (cgan)  {cGAN};
    \node [diagramEl, below = 1em of cgan] (hrmap)  {\scriptsize{HR Normal Map}};
    \node [diagramEl, below right = -4em and 1em of cgan] (final) {{\scriptsize{Rendering} \\ \includegraphics[trim={2cm 8cm 1.8cm 2.6cm},clip,height=8em]{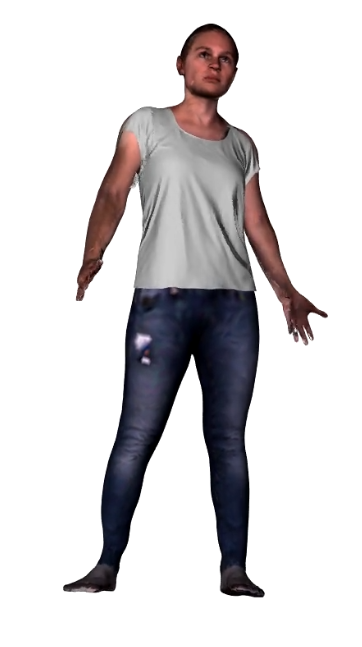}}};
    
    \draw[|-,-|,->, thick,] (estimation.east) to (reconstruction.west);
    \draw[|-,-|,->, thick] (statmodel.east) -|(9.8em,.5em)|- (reconstruction.west);
    \draw[|-,-|,->, thick,] (reconstruction.south) to (lrmap.north);
    \draw[|-,-|,->, thick,] (cgan.south) to (hrmap.north);
    \draw[|-,-|,->, thick] (lrmap.east) -|(18.7em,.5em)|- (cgan.west);
    \draw[|-,-|,->, thick,] (hrmap.east) to ($(final.west)+(0,-0.7)$);
    \draw[|-,-|,->, thick] (reconstruction.north) |-(25em,4.6em)-| (final.north);
    
    \draw[gray,thick,solid] ($(estimation.north west)+(-0.28,0.7)$) rectangle ($(final.south east)+(0.3,-0.1)$);
\end{tikzpicture}

%% file: images/cGAN.tikz
\tikzset{
  box/.style  = {draw,rectangle, minimum width=5cm, minimum height=1cm, text centered, text width=5cm, font=\Large},
  bigarrow/.style = {line width=.5mm, draw=blue!50!black!70!white, -triangle 60, postaction={draw, line width=1.5mm, shorten >=2mm, -}}
}

\begin{tikzpicture}[scale=0.9]
\linespread{0.7}
\node[inner sep=0pt,text width=.9cm,align=center] (input) at (-.8,1.7)
    {\includegraphics[width=\textwidth]{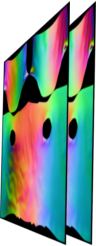}\\ \tiny{LR NM}\\ \tiny{$\{t, t-1\}$}};
\draw [->, line width=1.1] (-0.35,1.8) -- (0,1.8);

\draw [->, line width=1.1, style=dashed] (-0.35,1.4) -- (9,1.4);
    
\def\height{3.6}
\def\width{7.1}
	\draw[fill=white] (0,0) rectangle (\width,\height);
    \node[above] at (.5*\width,\height){Generator - U-Net};
    
    \def\layerw{0.2}
    \def\layerm{0.2}
    \def\layers{0.2}
    \def\heightdiff{0.35}
    \def\heightmax{0.2}
    \def\lowerm{0.5}
	\draw (\layers+0*\layerw+0*\layerm,0.1+\lowerm) rectangle (\layers+1*\layerw+0*\layerm,\height-\heightmax-0*\heightdiff);
    \draw (\layers+1*\layerw+1*\layerm,0.1+\lowerm) rectangle (\layers+2*\layerw+1*\layerm,\height-\heightmax-1*\heightdiff);
    \draw (\layers+2*\layerw+2*\layerm,0.1+\lowerm) rectangle (\layers+3*\layerw+2*\layerm,\height-\heightmax-2*\heightdiff);
    \draw (\layers+3*\layerw+3*\layerm,0.1+\lowerm) rectangle (\layers+4*\layerw+3*\layerm,\height-\heightmax-3*\heightdiff);
    \draw (\layers+4*\layerw+4*\layerm,0.1+\lowerm) rectangle (\layers+5*\layerw+4*\layerm,\height-\heightmax-4*\heightdiff);
    \draw (\layers+5*\layerw+5*\layerm,0.1+\lowerm) rectangle (\layers+6*\layerw+5*\layerm,\height-\heightmax-5*\heightdiff);
    \draw (\layers+6*\layerw+6*\layerm,0.1+\lowerm) rectangle (\layers+7*\layerw+6*\layerm,\height-\heightmax-6*\heightdiff);
    \draw (\layers+7*\layerw+7*\layerm,0.1+\lowerm) rectangle (\layers+8*\layerw+7*\layerm,\height-\heightmax-7*\heightdiff);
    
    \draw[fill=lightgray] (\layers+8*\layerw+8*\layerm+0.05,0.1+\lowerm) rectangle (\layers+9*\layerw+8*\layerm+0.05,\height-\heightmax-7.6*\heightdiff);
    
    \node[label={[label distance=0cm,text depth=-1ex,rotate=90]left:$256$}] at (\layers+0.15+0*\layerw+0*\layerm,0.7) {};
    \node[label={[label distance=0cm,text depth=-1ex,rotate=90]left:$128$}] at (\layers+0.15+1*\layerw+1*\layerm,0.7) {};
    \node[label={[label distance=0cm,text depth=-1ex,rotate=90]left:$64$}] at (\layers+0.15+2*\layerw+2*\layerm,0.6) {};
    \node[label={[label distance=0cm,text depth=-1ex,rotate=90]left:$32$}] at (\layers+0.15+3*\layerw+3*\layerm,0.6) {};
    \node[label={[label distance=0cm,text depth=-1ex,rotate=90]left:$16$}] at (\layers+0.15+4*\layerw+4*\layerm,0.6) {};
    \node[label={[label distance=0cm,text depth=-1ex,rotate=90]left:$8$}] at (\layers+0.15+5*\layerw+5*\layerm,0.5) {};
    \node[label={[label distance=0cm,text depth=-1ex,rotate=90]left:$4$}] at (\layers+0.15+6*\layerw+6*\layerm,0.5) {};
    \node[label={[label distance=0cm,text depth=-1ex,rotate=90]left:$2$}] at (\layers+0.15+7*\layerw+7*\layerm,0.5) {};
    
    \node[label={[label distance=0cm,text depth=-1ex,rotate=90]left:$1$}] at (\layers+0.15+8*\layerw+8*\layerm+0.05,0.5) {};
    
    \def\layerss{3.9}
    \draw (\layerss+6*\layerw+6*\layerm,0.1+\lowerm) rectangle (\layerss+7*\layerw+6*\layerm,\height-\heightmax-1*\heightdiff);
    \draw (\layerss+5*\layerw+5*\layerm,0.1+\lowerm) rectangle (\layerss+6*\layerw+5*\layerm,\height-\heightmax-2*\heightdiff);
    \draw (\layerss+4*\layerw+4*\layerm,0.1+\lowerm) rectangle (\layerss+5*\layerw+4*\layerm,\height-\heightmax-3*\heightdiff);
    \draw (\layerss+3*\layerw+3*\layerm,0.1+\lowerm) rectangle (\layerss+4*\layerw+3*\layerm,\height-\heightmax-4*\heightdiff);
    \draw (\layerss+2*\layerw+2*\layerm,0.1+\lowerm) rectangle (\layerss+3*\layerw+2*\layerm,\height-\heightmax-5*\heightdiff);
    \draw (\layerss+1*\layerw+1*\layerm,0.1+\lowerm) rectangle (\layerss+2*\layerw+1*\layerm,\height-\heightmax-6*\heightdiff);
    \draw (\layerss+0*\layerw+0*\layerm,0.1+\lowerm) rectangle (\layerss+1*\layerw+0*\layerm,\height-\heightmax-7*\heightdiff);
    
    \def\margmir{0.08}
    \draw[fill=lightgray] (\layerss+7*\layerw+7*\layerm+\margmir,0.1+\lowerm+\margmir) rectangle (\layerss+8*\layerw+7*\layerm+\margmir,\height-\heightmax-0*\heightdiff+\margmir);
    \draw[fill=lightgray] (\layerss+6*\layerw+6*\layerm+\margmir,0.1+\lowerm+\margmir) rectangle (\layerss+7*\layerw+6*\layerm+\margmir,\height-\heightmax-1*\heightdiff+\margmir);
    \draw[fill=lightgray] (\layerss+5*\layerw+5*\layerm+\margmir,0.1+\lowerm+\margmir) rectangle (\layerss+6*\layerw+5*\layerm+\margmir,\height-\heightmax-2*\heightdiff+\margmir);
    \draw[fill=lightgray] (\layerss+4*\layerw+4*\layerm+\margmir,0.1+\lowerm+\margmir) rectangle (\layerss+5*\layerw+4*\layerm+\margmir,\height-\heightmax-3*\heightdiff+\margmir);
    \draw[fill=lightgray] (\layerss+3*\layerw+3*\layerm+\margmir,0.1+\lowerm+\margmir) rectangle (\layerss+4*\layerw+3*\layerm+\margmir,\height-\heightmax-4*\heightdiff+\margmir);
    \draw[fill=lightgray] (\layerss+2*\layerw+2*\layerm+\margmir,0.1+\lowerm+\margmir) rectangle (\layerss+3*\layerw+2*\layerm+\margmir,\height-\heightmax-5*\heightdiff+\margmir);
    \draw[fill=lightgray] (\layerss+1*\layerw+1*\layerm+\margmir,0.1+\lowerm+\margmir) rectangle (\layerss+2*\layerw+1*\layerm+\margmir,\height-\heightmax-6*\heightdiff+\margmir);
    \draw[fill=lightgray] (\layerss+0*\layerw+0*\layerm+\margmir,0.1+\lowerm+\margmir) rectangle (\layerss+1*\layerw+0*\layerm+\margmir,\height-\heightmax-7*\heightdiff+\margmir);
    
    \node[label={[label distance=0cm,text depth=-1ex,rotate=90]left:$256$}] at (\layerss+0.15+7*\layerw+7*\layerm,0.7) {};
    \node[label={[label distance=0cm,text depth=-1ex,rotate=90]left:$128$}] at (\layerss+0.15+6*\layerw+6*\layerm,0.7) {};
    \node[label={[label distance=0cm,text depth=-1ex,rotate=90]left:$64$}] at (\layerss+0.15+5*\layerw+5*\layerm,0.6) {};
    \node[label={[label distance=0cm,text depth=-1ex,rotate=90]left:$32$}] at (\layerss+0.15+4*\layerw+4*\layerm,0.6) {};
    \node[label={[label distance=0cm,text depth=-1ex,rotate=90]left:$16$}] at (\layerss+0.15+3*\layerw+3*\layerm,0.6) {};
    \node[label={[label distance=0cm,text depth=-1ex,rotate=90]left:$8$}] at (\layerss+0.15+2*\layerw+2*\layerm,0.5) {};
    \node[label={[label distance=0cm,text depth=-1ex,rotate=90]left:$4$}] at (\layerss+0.15+1*\layerw+1*\layerm,0.5) {};
    \node[label={[label distance=0cm,text depth=-1ex,rotate=90]left:$2$}] at (\layerss+0.15+0*\layerw+0*\layerm,0.5) {};
    
    \draw [->, color=red!50!black, line width=1.1] (\layers+2*\layerw+1*\layerm+0.05,\height-\heightmax-1*\heightdiff-0.15) -- (\layerss+6*\layerw+7*\layerm-0.2,\height-\heightmax-1*\heightdiff-0.15);
    \draw [->, color=red!50!black, line width=1.1] (\layers+3*\layerw+2*\layerm+0.05,\height-\heightmax-2*\heightdiff-0.15) -- (\layerss+5*\layerw+6*\layerm-0.2,\height-\heightmax-2*\heightdiff-0.15);
    \draw [->, color=red!50!black, line width=1.1] (\layers+4*\layerw+3*\layerm+0.05,\height-\heightmax-3*\heightdiff-0.15) -- (\layerss+4*\layerw+5*\layerm-0.2,\height-\heightmax-3*\heightdiff-0.15);
    \draw [->, color=red!50!black, line width=1.1] (\layers+5*\layerw+4*\layerm+0.05,\height-\heightmax-4*\heightdiff-0.15) -- (\layerss+3*\layerw+4*\layerm-0.2,\height-\heightmax-4*\heightdiff-0.15);
    \draw [->, color=red!50!black, line width=1.1] (\layers+6*\layerw+5*\layerm+0.05,\height-\heightmax-5*\heightdiff-0.15) -- (\layerss+2*\layerw+3*\layerm-0.2,\height-\heightmax-5*\heightdiff-0.15);
    \draw [->, color=red!50!black, line width=1.1] (\layers+7*\layerw+6*\layerm+0.05,\height-\heightmax-6*\heightdiff-0.15) -- (\layerss+1*\layerw+2*\layerm-0.2,\height-\heightmax-6*\heightdiff-0.15);
    \draw [->, color=red!50!black, line width=1.1] (\layers+8*\layerw+7*\layerm+0.05,\height-\heightmax-7*\heightdiff-0.15) -- (\layerss+0*\layerw+1*\layerm-0.2,\height-\heightmax-7*\heightdiff-0.15);

\node[inner sep=0pt,text width=.9cm,align=center] (input) at (\width+.7,1.7)
    {\includegraphics[width=.7\textwidth]{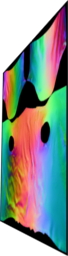}\\ \tiny{HR NM}\\ \tiny{$\{ t \}$}};
\draw [->, line width=1.1] (\width,1.8) -- (\width+0.35,1.8);
    
\def\startd{9}
\def\widthd{2.8}
    \draw (\startd,0.5) rectangle (\startd+\widthd,\height);
    \node[above] at (\startd+0.5*\widthd,\height){Discriminator};
    \node[above] at (\startd+0.5*\widthd,\height-0.5){\tiny{PatchGAN}};
    
    \node[inner sep=0pt,opacity=0.3] (nm) at (\startd+1.4,1.9)
    {\includegraphics[width=.18\textwidth]{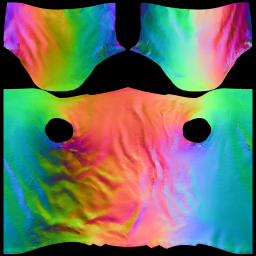}};
    \node[inner sep=0pt] (patch) at (\startd+1.73,1.74)
    {\includegraphics[width=.047\textwidth,trim={3.5cm, 2cm, 1.5cm, 3cm},clip]{output.png}};
    \draw (\startd+1.44,1.45) rectangle (\startd+2.02,2.05);
    \node[above] at (\startd+1.44+0.29,1.1){\tiny$70 \times 70$};
    
    \draw [->] (\startd+1.5,0.5) |- (\startd+2,0.3);
    \node[above] at (\startd+2.35,0){$0,1$};
    
	\draw [->, line width=1.1] (\width+1.02,1.8) -- (\startd,1.8);
    \draw [->] (\startd+0.8,0.3) -| (\startd+1.3,0.5);
    \node[above] at (\startd+0.6,0.03){\emph{gt}};

\end{tikzpicture}